\documentclass[runningheads]{llncs}
\usepackage{graphicx}
\usepackage{times}
\usepackage{latexsym}
\usepackage{bm}

\usepackage{cite}
\usepackage{amsmath,amssymb,amsfonts}

\usepackage{textcomp}

\usepackage{microtype}

\usepackage{enumerate}
\usepackage{booktabs}
\usepackage{makecell}
\usepackage{wrapfig}
\usepackage{picins}

\usepackage{subfig}
\usepackage{algorithmicx}
\usepackage{float}
\usepackage{framed} 
\usepackage[marginal,multiple]{footmisc}

\usepackage{times}
\usepackage[figuresright]{rotating}

\usepackage[T1]{fontenc}

\usepackage[utf8]{inputenc}

\usepackage{calligra}
\usepackage{algpseudocode}

\usepackage{xspace}
\usepackage{moreverb}
\usepackage{fontenc}
\usepackage{fancybox}
\usepackage{color}
\usepackage{xcolor}
\usepackage{colortbl}
\usepackage{array}
\usepackage{multirow}
\usepackage{multicol}
\usepackage{listings}
\usepackage{graphicx}
\usepackage{setspace}
\usepackage{wasysym}
\usepackage{balance}
\usepackage{csvsimple}
\usepackage{longtable}
\usepackage[skip=0pt,labelfont=bf]{caption}
\usepackage{url}
\usepackage{lscape}
\usepackage{rotating}
\usepackage{tikz}
\usepackage[normalem]{ulem}
\usepackage{enumitem}

\usepackage[most]{tcolorbox}

\usepackage{threeparttable}
\usepackage{marvosym}
\usepackage{pifont}
\usepackage{courier}
\usepackage{afterpage,lipsum}
\usepackage{mathrsfs}
\usepackage{dsfont}
\usepackage{calligra}
\usepackage{algpseudocode}

\algnewcommand\algorithmicforeach{\textbf{for each}}
\algdef{S}[FOR]{ForEach}[1]{\algorithmicforeach\ #1\ \algorithmicdo}

\newcolumntype{L}[1]{>{\raggedright\let\newline\\\arraybackslash\hspace{0pt}}m{#1}}
\newcolumntype{C}[1]{>{\centering\let\newline\\\arraybackslash\hspace{0pt}}m{#1}}
\newcolumntype{R}[1]{>{\raggedleft\let\newline\\\arraybackslash\hspace{0pt}}m{#1}}

\definecolor{codegreen}{rgb}{0,0.6,0}
\definecolor{codered}{rgb}{1,0,0}
\definecolor{codegray}{rgb}{0.5,0.5,0.5}
\definecolor{codepurple}{rgb}{0.58,0,0.82}
\definecolor{backcolour}{rgb}{0.95,0.95,0.92}
\definecolor{lightgray}{gray}{0.9}
\definecolor{cadmiumgreen}{rgb}{0.0, 0.42, 0.24}
\definecolor{caribbeangreen}{rgb}{0.0, 0.8, 0.6}

\lstdefinestyle{mystyle}{
    commentstyle=\color{codegreen},
    keywordstyle=\color{magenta},
    numberstyle=\small\color{black},
    stringstyle=\color{codepurple},
    basicstyle=\scriptsize\ttfamily,
    breakatwhitespace=false,
    breaklines=true,
    captionpos=b,
    keepspaces=true,
    showspaces=false,
    showstringspaces=false,
    showtabs=false,
    tabsize=2
}

\lstdefinelanguage{diff}{
  morecomment=[f][\color{blue}]{@@},     
  morecomment=[f][\color{red}]-,         
  morecomment=[f][\color{codegreen}]+,       
  morecomment=[f][\color{red}]{---}, 
  morecomment=[f][\color{codegreen}]{+++}
}

\lstdefinelanguage{text}{
  breaklines=false
}

\setlist{noitemsep} 

\definecolor{darkpastelred}{rgb}{0.76, 0.23, 0.13}
\definecolor{ao(english)}{rgb}{0.0, 0.5, 0.0}

\definecolor{darkpastelred}{rgb}{0.76, 0.23, 0.13}
\definecolor{ao(english)}{rgb}{0.0, 0.5, 0.0}

\definecolor{yellow}{RGB}{255,255,153}
\definecolor{grey}{RGB}{224,224,224}

\newboolean{showcomments}
\setboolean{showcomments}{true}
\ifthenelse{\boolean{showcomments}}
 { \newcommand{\mynote}[2]{
      \fbox{\bfseries\sffamily\scriptsize#1}
        {\small$\blacktriangleright$\textsf{\emph{#2}}$\blacktriangleleft$}}}
        { \newcommand{\mynote}[2]{}}

\definecolor{DarkOrange}{rgb}{0.8,0.3,0.0}
\definecolor{DarkCyan}{rgb}{0.0, 0.55, 0.55}
\definecolor{DarkCyel}{rgb}{1.0, 0.49, 0.0}
\definecolor{yellow-green}{rgb}{0.6, 0.8, 0.2}

\newcolumntype{?}{!{\vrule width 1pt}}

\newcommand{\toolname}{GE-Blender\xspace} 
\newcommand{\datasetname}{Wizard of Wikipedia\xspace}
\newcommand{\moduleA}{Graph Knowledge Enhancement\xspace}
\newcommand{\moduleB}{Named Entity Tags Enhancement\xspace}
\begin{document}
\title{\toolname: Graph Based Knowledge Enhancement for Blender}
%
%

\author{Xiaolei Lian\inst{1*} \and
Xunzhu Tang\inst{2*} \and
Yue Wang\inst{3}}

\institute{Hebei Normal University, China,
\email{lian19931201@gmail.com}\\ \and
University of Luxembourg, Luxembourg,
\email{xunzhu.tang@uni.lu}\\ \and
Haomo Technology Co., Ltd., China,
\email{wangyue1@haomo.ai}}

\maketitle              
%
\begin{abstract}
Although the great success of open-domain dialogue generation, unseen entities can have a large impact on the dialogue generation task. It leads to performance degradation of the model in the dialog generation. Previous researches used retrieved knowledge of seen entities as the auxiliary data to enhance the representation of the model. Nevertheless, logical explanation of unseen entities remains unexplored, such as possible co-occurrence or semantically similar words of them and their entity category. In this work, we propose an approach to address the challenge above. We construct a graph by extracting entity nodes in them, enhancing the representation of the context of the unseen entity with the entity's 1-hop surrounding nodes. Furthermore, We added the named entity tag prediction task to apply the problem that the unseen entity does not exist in the graph. We conduct our experiments on an open dataset \datasetname and the empirical results indicate that our approach outperforms the state-of-the-art approaches on \datasetname.
\end{abstract}

\section{Introduction}

Unseen entity problem leads to the failure of the model in recognizing semantics in open-domain dialogue generation\cite{cui2021knowledge,peng2022control,peng2021aper,peng2022you,wang2022automatically} and some previous methods improved on this issue.
and some previous methods have made many improvements to this issue.
Knowled-GPT uses reinforcement learning for knowledge selection\cite{zhao2020knowledge,tang2021ckg,wang2022hienet} and feed knowledge as input of the model.
Knowledge Grounded Blender focuses on fixing the problem of how to supply the knowledge of seen entities\cite{zhao2020knowledge}. It addresses this problem in a good performance, but ignores the cases of unseen entities. 
Knowledge Enhanced Blender(KE-Blender) aims to enhance the representation of context with seen entities' knowledge\cite{cui2021knowledge}. It forces the model to generate the definition and hypernym of seen entities, then updates model parameters for context representation enhancement. KE-Blender gives a state-of-the-art performance on \datasetname. But if we replace the currently seen entity with a semantically different unseen entity and keep the context unchanged, the context representation enhancement of KE-Blender will not work well. 

\begin{figure}[htbp]
\centering
\begin{minipage}[t]{0.48\textwidth}
\centering
\includegraphics[width=6cm]{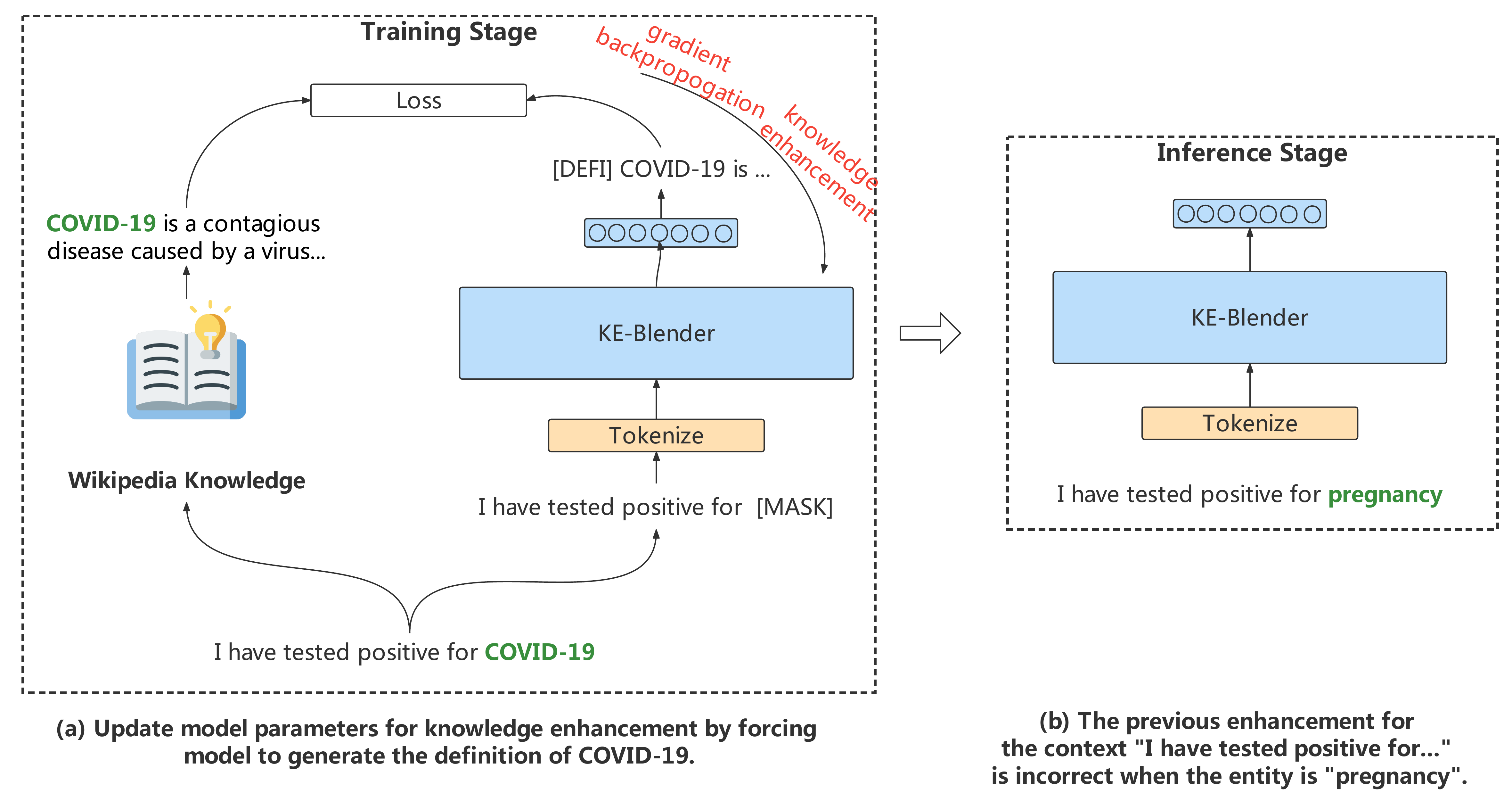}
\caption{An illustration for wrong enhancement in the inference stage of KE-Blender.}
\label{fig:badcase_covid}
\end{minipage}
\begin{minipage}[t]{0.48\textwidth}
\centering
\includegraphics[width=6cm]{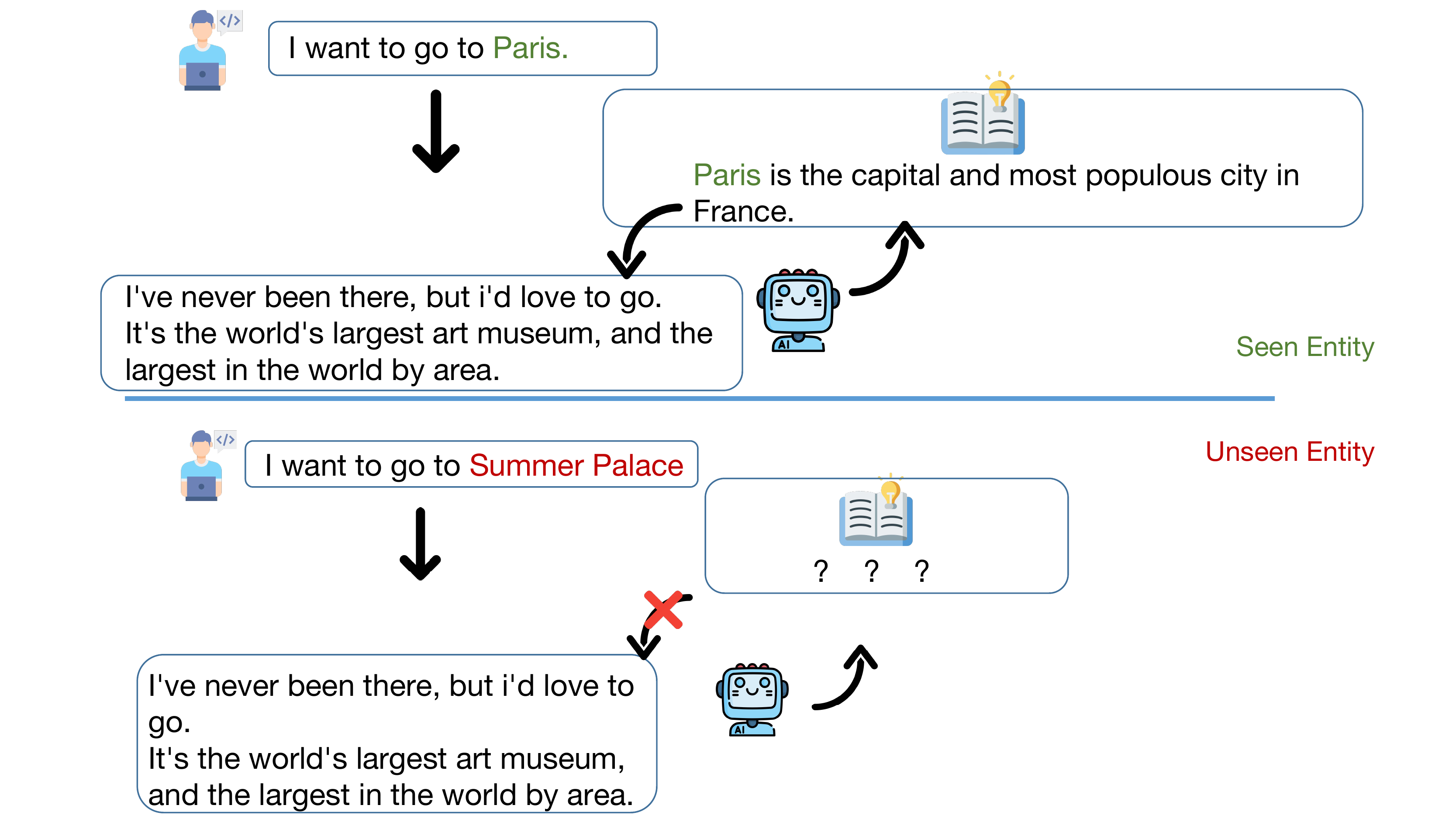}
\caption{The generation result of KE-Blender on seen entity scenario and unseen entity respectively}
\label{fig:badcase_paris}
\end{minipage}
\end{figure}

Describe with an example in Figure \ref{fig:badcase_covid}, given the sentence "\textit{I have tested positive for COVID-19}" and the definition of "\textit{COVID-19}",  the model will generate the response "\textit{I'm sorry to hear that.}" But if we replace the entity "\textit{COVID-19}" to "\textit{pregnancy}". The previous enhancement about the context "\textit{I have tested positive}" with the definition of "\textit{COVID-19}" will not positively impact the result.
Take another proven example as the illustration in Figure \ref{fig:badcase_paris},
the KE-Blender doesn't work well when there are some unseen entity in the input sentence.

In order to solve this problem, we propose a novel method that enhances knowledge representation with graph knowledge and named entity recognition information in the dialogue generation task. We call our method as \textbf{G}raph Based Knowledge \textbf{E}nhancement for \textbf{Blender}(\textbf{\toolname}).

Our approach have two modules: 
(1) \textbf{\moduleA}: We integrate the 1-hop node of the entity in the document searched from the graph constructed with dialog and retrieved knowledge into the dialogue generation task as knowledge enhancement;
(2) \textbf{\moduleB}: For an entity that does not exist in the graph, we replace the entity with the entity tag generated by Named Entity Recognition(NER)\cite{lample2016neural} model and use the sequence replaced by tags as the target sequence to make the model generate.

Specifically, in the \moduleA, we first build a graph with dialogue data and retrieved knowledge in the dialogue generation dataset(such as \datasetname). Secondly, we extract 1-hop nodes of proper nouns in dialogue and flatten these nodes as an ordered word sequence by the edge weights. Finally, we regard the word sequence as the target sequence and make the model generate the sequence to enhance the context representation. 
In this way, we can use the global co-occurrence information of entities to enhance the representation of the context of entities.
In the \moduleB, we use the NER model to predict the tags of named entities and replace named entities with their NER tags as a target sequence. Then we force the model to generate the replaced sequence. Through this task, the context representation can be further enhanced.

The contributions of this paper are as follows:

\begin{itemize} [leftmargin=*]
    \item We use graph knowledge to enhance the representation of the context of the entity, allowing the dialogue generation model to refer to global word co-occurrence information for solving the unseen entity problem.
    \item We use NER tags to guide the model to enhance the representation of the context of the entity.
\end{itemize}

Results on the \datasetname benchmark show that the proposed model brings performance improvement. The proposed method achieves \textbf{16.7} and \textbf{15.3} F1-score on \datasetname Test Seen and Test Unseen and PPL achieves \textbf{16.5} and \textbf{20.0} respectively. We release our code and dataset at https://github.com/Fakekid/GE-Blender.

\section{Related Work}

Here we review the related works, including open-domain dialogue generation, knowledge enhancement approaches and unseen entity problems.

\subsection{Open-domain Dialogue Generation}

As the definition of open-domain dialogue generation in \cite{kann2022open}, the system must output a fluent, engaging, and meaningful natural language response based on the given previous dialogue turns. Different from task-based dialogue generation, open-domain dialogue generation does not need to have a clear goal, and the content of the interlocutor is often very casual hence open-domain dialogue will appear with unseen entities.

A common base for open-domain dialogue generation model is Blender: a standard Seq2Seq Transformer architecture with pre-training~\cite{wang2022hienet,wang2021large,liu2019roberta,tang2023app}. Some recent research works use it as a basebone.

\subsection{Knowledge Enhanced for Dialogue System} 

The knowledge enhancement method is relatively popular in recent years, and it has been applied in various fields, such as language model\cite{zhang2019ernie}, text classification model\cite{chen2019deep}, chat-bot model\cite{cui2021knowledge}, etc. The knowledge enhancement method is to integrate the knowledge mastered by humans into the input data to allow the model to learn and enhance the representation ability of the model. Taking ERNIE\cite{zhang2019ernie} as an example, it extracts the entities in the input sentence, performs multi-head attention encoding, and fuses them with the hidden state of the token corresponding to the entity encoded by the default multi-head attention. For the text classification sense, STCKA\cite{chen2019deep} first extracts the concept of the entity in the input sentence from a knowledge base, then run the knowledge encoder layer. Finally, fuses the knowledge encoding and default multi-head attention output.

In the open-domain dialogue generation task, we integrate the 1-hop nodes of the entity in the graph as enhanced knowledge and integrate it into the input, so that the model can pay attention to the global co-occurrence words of the entity.

\subsection{Unseen Entity Problems} 

The unseen entity problem is a common problem in open-domain dialogue generation. What topics we talk about often are new and not in the dataset that the pre-trained model used. Some previous methods do not fix this problem, they didn't work well on unseen entities. The KG-Blender\cite{zhao2020knowledge} concatenates the dialogue sentence and the seen entity's definition in retrieved knowledge as a knowledge-enhanced method. But this approach does not work when the entity does not appear in the training stage. The KE-Blender\cite{cui2021knowledge} uses two tasks for knowledge enhancement, one is to generate the definition of the seen entity in retrieved knowledge, and the other is to generate the hypernym of the seen entity. Both of these tasks can enhance the representation of the model, but when an unseen entity has no hypernym and definition, the KE-Blender also does not work well.

For these hard problems, we use graph knowledge to enhance the representation of the dialogue generation model in the unseen entities sense.

\section{Approach}

\subsection{Base Model}

\noindent \textbf{Named Entity Recognition}  Named Entity Recognition (NER) aims at extracting and classifying mentions of rigid designators, from text, such as proper names, biological species, and temporal expressions \cite{nadeau2007survey}. 
NER not only acts as a standalone tool for information extraction but also plays an essential role in a variety of natural language processing (NLP) applications such as text understanding \cite{li2020survey}. Therefore, the NER model can usually capture context information to complete accurate prediction tasks. In this regard, we can also use the characteristics of the NER task to strengthen the open-domain dialog generation model.

The task of NER can be expressed by the following formula:
\begin{equation}
\label{eq:ner_forward}
p(y_{i}|x_{i},y_{:i-1};\theta ) = \sigma(NER(x_{i},y_{:i-1};\theta)) \in \mathbb{R}^{pl}
\end{equation}

Where $ x_i $ is the i th token of input sentence, and $ y_i $ is the i th named entity tag generated by NER model.

We use Bi-LSTM as our base NER model and use the AllenNLP\cite{Gardner2017AllenNLP} library to complete the NER task.

\noindent \textbf{Blender}  We use Blender\cite{roller2020recipes} as our basebone, Blender is a pre-trained dialogue generation model based on sequence-to-sequence transformer structure. It uses an encoder to encode the input sequence and uses the decoder to generate a decoded sequence. 

Given a complete dialogue $ U = \{u_1, ..., u_{l-1} \} $, we concatenate all previous sequences of each response $ U_{:i-1} = \{x_1, x_2, . . . , x_T \} $ as input sequence and the current reply $ R = \{y1, y2, . . . , y_{T^{'}} \} $ as the target sequence. The model first encodes the input sequence with the encoder:

\begin{equation}
h^{enc} = \text{TRM\_ENC}(U) \in \mathbb{R}^{pl}
\end{equation} 

and then decodes the hidden states based on the encoder to generate the response sequence:

\begin{equation}
h^{dec}_t = \text{TRM\_DEC}(h_{enc}, y_{1:t-1}) \in \mathbb{R}^{pl}
\end{equation}

 Where the $p$, $l$ is the length of vocabulary and the sentence length.

In the $t$ th step of the decoder, $h_{enc}$ and previous output tokens $y_{1:t-1}$ are then as inputs, yielding a representation using attention layers\cite{vaswani2017attention}.

The generative probability distribution of $y_t$ is given by
\begin{equation}
p(y_t|U, y_{1:t-1}) = softmax(W_o h^{dec}_t + b_o) \in \mathbb{R}^{p}
\end{equation}

where $W_o$ and $b_o$ are trainable parameters.

We adopt Blender-90M \cite{roller2020recipes} to initialize our Seq2Seq Transformer model, which has been pre-trained on 1.5B training examples from Reddit 2019.


\subsection{Our Architecture}

\begin{figure*}[htp]
    \centering
    \includegraphics[width=13cm]{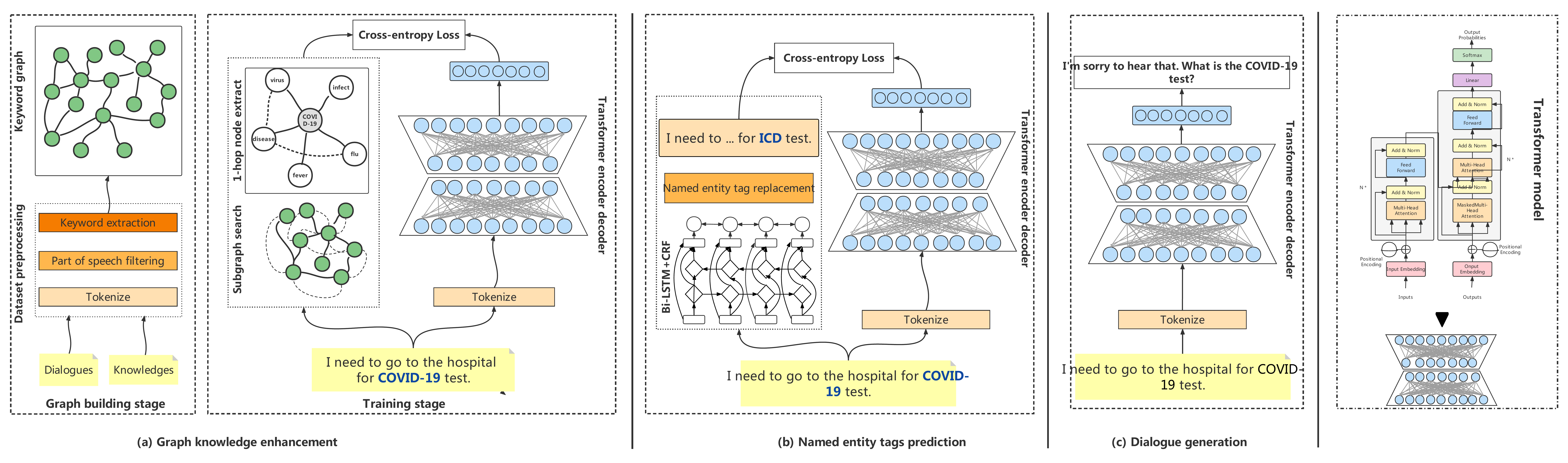}
    \caption{The architecture of \toolname. (a) illustrate the graph construction and model train progress of \moduleA; (b) shows how the \moduleB works. It makes a target sequence with the Bi-LSTM based model; (c) is a standard structure of Blender, it contains encoder and decoder both based on transformer.}
    \label{fig:architecture}
\end{figure*}

\noindent \textbf{\moduleA}  As an illustration of Figure \ref{fig:architecture} (a), we first convert each document in the dialogue and the retrieval knowledge in the \datasetname to a keywords sequence, which can extract the more important words in the sentence, thus ignoring the unimportant content. In the next step, we build a graph with keywords sequences, where each word is a node, and there will be an edge between the two nodes if these nodes appear in a document together.\cite{nikolentzos2020k}

At the model training stage, we extract the 1-hop nodes of the proper nouns in a document from the graph and sort the 1-hop nodes according to the edge weight. Continually, we take the top K nodes as a target sequence to make the model generate it to enhance the context representation.

\begin{equation}
\mathcal{N}(v) = \sum_{u \in \mathcal{U}}{\mathds{1}[(v, u) \in \mathcal{E}]}
\end{equation}

The $\mathcal{N}(*)$ is a function of getting the neighborhood vertexes. The $\mathcal{U}$, and $\mathcal{E}$ are vertex set and edge set on the graph respectively. And the 1-hop sequence is:

\begin{equation}
seq(v) = topk(\mathop{argsort}\limits_{u}(\{w_{v, u} | u \in \mathcal{N}(v)\}))
\end{equation}

where $w_{v,u}$ is the edge weight between vertex $v$ and $u$:

\begin{equation}
w_{v, u_i} = \frac{exp(\#(v,u_i))}{\sum_k^K exp(\#(v, u_k))}
\end{equation}

The function $\#(*)$ is the statistical function of the number of co-occurrences.

\begin{figure}[htp]
    \centering
    \includegraphics[width=8cm]{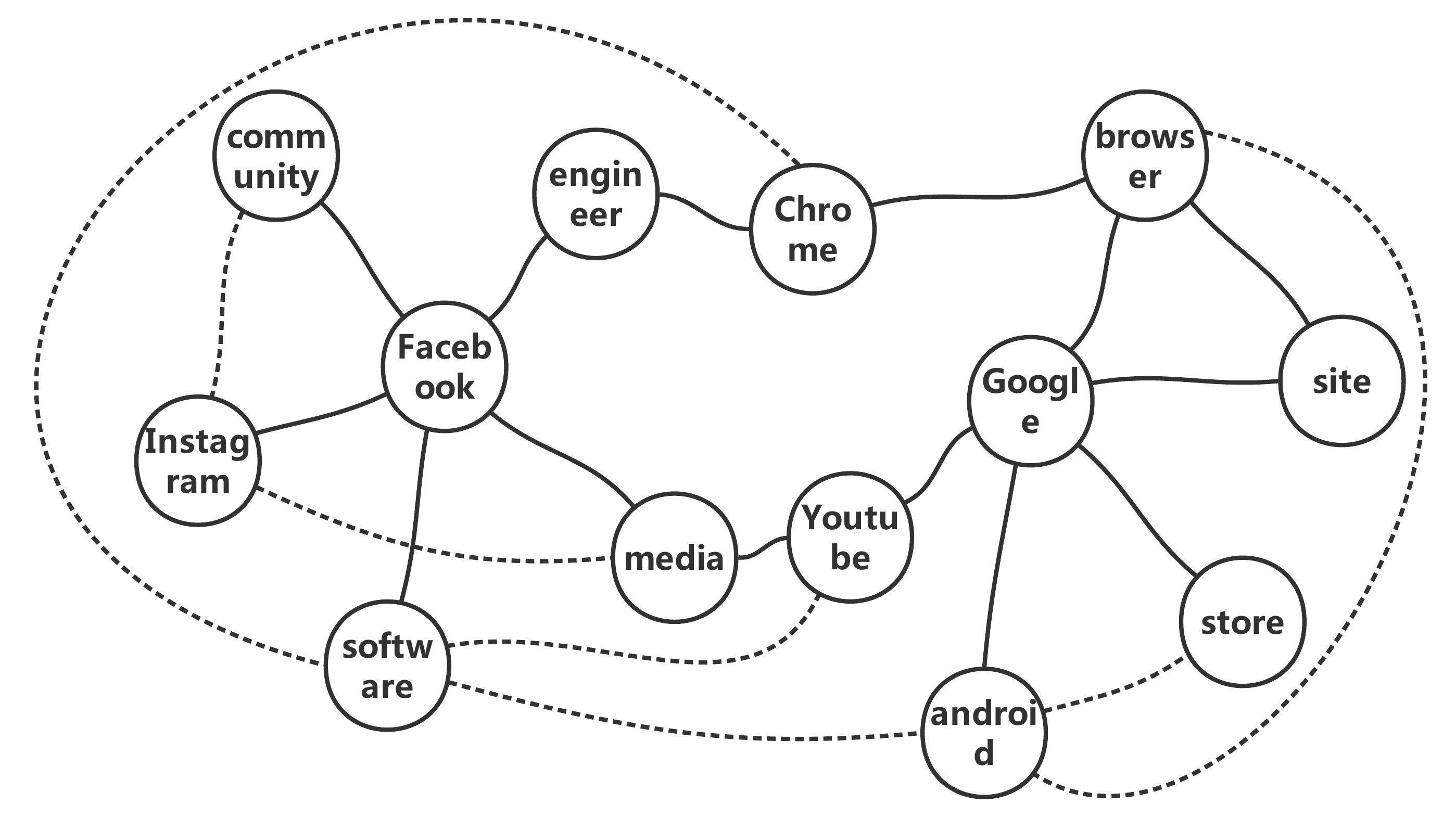}
    \caption{The 1-hop subgraph for entities "Facebook" and "Google".}
    \label{fig:one_hop_node}
\end{figure}

Like the Figure \ref{fig:one_hop_node} illustration, the entity "Facebook" is most related with "social", "media", "picture" and so on. And "Google" is most related with "retrieve", "email", "advertise". 
We use a specific start token "[GRAPH]" to mark the current task is node sequence generation. The loss of this module is as follows:

\begin{equation}
\mathcal{L}_{graph} = -\sum_{t=1}^{|seq(v)|}log(p(y_t|x_{t}, \mathbf{h}))
\end{equation}

where the $y_t$ is the top-k 1-hop sequence item when the $x_t$ is a noun or proper noun.

\noindent \textbf{\moduleB}  For entities that are not in the graph, we let the model predict its named entity tags \cite{lample2016neural}, thus performing knowledge enhancement on nodes that are not in the graph.

Intuitively, the correct recognition of named entity tags indicates the model to understand the context of the entity. We replace the entities in the dialogue with their named entity tags predicted by the NER model. In the second step, we take the replaced sequence as the target sequence with a start token "[NER]" and then train the model. The loss of this module is as follows:

\begin{equation}
\mathcal{L}_{ner} = -\sum_{t=1}^{T}log(p(z_t|x_{t}, \mathbf{h}_{1:t-1}))
\end{equation}

where the $z_t$ is the named entity tag when the $x_t$ is a named entity, otherwise, it is a token same as the origin sequence.

The part (b) of Figure \ref{fig:architecture} shows the progress of the \moduleB. We make an origin dialogue sentence as the input, then we use the NER model to obtain the named entity tags, find the named entities in the original sequence and replace them with named entity tags then train the model.


\noindent \textbf{Training}  In addition to optimizing the loss of dialog generation tasks, we also optimize the loss of \moduleA and \moduleB:

\begin{equation}
\mathcal{L}_{total} = \sum_{i=1}^{|O|}\mathcal{L}_{dialogue} + \sum_{i=1}^{|P|}\mathcal{L}_{graph} + \sum_{i=1}^{|Q|}\mathcal{L}_{ner}
\end{equation}

where $|O|$, $|P|$ and $|Q|$ represent the number of training instances of the three task respectively.


\noindent \textbf{Inference}  During inference, we only take the dialogue sentence as the input sequence and extract the 1-hop nodes of proper nouns in the sentence from the graph as the auxiliary data. We don't need any knowledge about entities. At the sequence generation stage, we adopt greedy search to select the highest probability token at each time step\cite{cui2021knowledge}.

\section{Experimental Setup}
\label{sec: exp}

We first list some research questions to prove the effectiveness of our method. Then, we describe the dataset used for responding to the questions. Finally, we present the evaluation metrics used in our study.

\subsection{Research Questions}

To validate the proposed approach, we define four main research questions including What’s the influence of unseen entities?, Overall Performance of \toolname, and Case study:

\begin{itemize}[leftmargin=*]
    \item {\bf RQ-1:} What’s the influence of unseen entities? 
    
    Compared to the dialogue generation on Seen Entity, the dialogue generation on Unseen Entity will be more difficult. We sample some cases to prove it.

    We use Bilingual Evaluation Understudy(BLEU) and unigram F1-score(F1) metrics to demonstrate that unseen entities affect the results of dialogue generation. 

	\item {\bf RQ-2:} What is the improvement of \toolname against the state-of-the-art?

	To resolve the unseen entities problem, we try to add the information related to the unseen entities to the model to optimize the model prediction results. We will use some metrics to show the improvement brought by our model.
	
	We can prove that our method has advantages on Unseen Entity compared to the current state-of-the-art. We compare our approach with transformer-based dialogue generation models and Blender series models.
    
    {\bf SKT + GPT-2} \cite{zhao2020knowledge} uses the pre-trained transformer based model SKT to select knowledge and feeds the knowledge to dialogue generation model.
    
    {\bf SKT + PIPM + KDBTS} \cite{chen-etal-2020-bridging} uses the posterior knowledge selection and distillation module to assist the prior knowledge selection module in knowledge selection.
    
    {\bf KnowledGPT} \cite{zhao2020knowledge} adopts reinforcement learning to optimize the BERT based knowledge selection module and input the selected knowledge and original data into the model. It gets the most advanced performance on the Wizard.
    
    {\bf Blender-FT} Blender is a large-scale dialogue pre-training model. We fine-tune the Blender on \datasetname training set without utilizing external knowledge.
    
    {\bf KG-Blender} KG-Blender concatenates the context and the knowledge as input for knowledge enhancement. We fine-tune the KG-Blender model on \datasetname
    
    {\bf KE-Blender} KE-Blender adds definition generation and hypernym generation tasks, trying to let the model grasp the knowledge of the context of the entity so that additional knowledge is not used as input during inference. We train it with default hyperparameters.
    	
	\item {\bf RQ-3:} How does our \toolname perform in real cases?
	
	We extract some cases to test the learning efficiency of the proposed approach. At the same time, the output results of other methods are compared with ours.

\end{itemize}

\subsection{Parameter Settings}

We implement our architecture with a transformer encoder-decoder model and use the structure and pre-training parameters of blenderbot-90M. We trained our model on two 3090Ti GPUs and set the optimizer to AdamW. The initial learning rate is 1e-5 whereas the batch size is 16. The 512 is the maximum length of the sequence in the model. When the length of a sequence is more than 512, we will truncate it.

\subsection{Dataset \& Evaluation Metric}

We use the Wizard of Wikipedia\cite{dinan2018wizard} as our dataset. It is a chit-chat dialogue benchmark. There is some retrieved knowledge from the Wikipedia documents in each utterance for reference. The dataset contains 18,430 training dialogues and 1933 test dialogues. For the test data, it is split into Test Seen and Test Unseen based on whether the topic is appear in the training set\cite{cui2021knowledge}. There are 965 dialogues in Test Seen and the other 968 dialogues in Test Unseen. In addition, we also distinguish between "with knowledge during inference" and "without knowledge during inference" based on whether the retrieved knowledge is spliced at the end of the input sentence during inference.

For fair comparison with baselines in different RQ, we unitize following metrics: PPL and F1-score. We also use ROUGE as a secondary evaluation metric.

\textbf{PPL} \cite{dinan2018wizard}, \cite{Kim2020SequentialLK} and \cite{cui2021knowledge} use the perplexity of the groundtruth response (PPL) and unigram F1-score (F1) as majority metrics for dialogue generation model. Perplexity (PPL) is one of the most common metrics for evaluating language models. Given a model and an input text sequence, perplexity measures how likely the model is to generate the input text sequence\cite{jelinek1977perplexity}. Therefore, we can evaluate our model with PPL. Generally, the calculation formula of ppl is:

\begin{equation}
    \text{PPL}(X) = \exp\{ -\frac{1}{t}\sum_{i}^{t}\log p_{\theta}(x_i|x_{<i}) \}
\end{equation}

where $X=\{x_0,x_1, ...,x_t\}$ is a input sequence of length $t$.

\textbf{ROUGE-N}\cite{lin2004rouge} is Recall-Oriented Understudy for Gisting Evaluation. It includes measures to automatically determine the quality of a summary by comparing it to other (ideal) summaries created by humans. ROUGE-N is computed as follows:

\begin{small}
\begin{equation}
\begin{aligned}
\text{ROUGE-N} &= \frac{\sum\limits_{S \in \{Ref\_Sum\}} \sum\limits_{gram_n \in S} \mathop{Count}\limits_{match}(gram_n)}{\sum\limits_{S \in \{Ref\_Sum\}} \sum\limits_{gram_n \in S} Count(gram_n)}
\end{aligned}
\end{equation}
\end{small}
where {\it Ref} is short for reference and {\it Sum} means summary.

\section{Experiments And Results}
\label{sec:eval}

\subsection{{\bf [RQ-1]:} Influence of unseen entity}
\label{subsec:rq1}

Our experiments show that the unseen entity restricts the the performance of the model. Blender generates the next round of dialogue only based on the content of the previous dialogue. The KG-Blender concatenates the dialogue and retrieved knowledge as the model's input. If we have some unseen entities, it also doesn't work well. If we switch the data from Test Seen to Test Unseen, these approaches will show significant performance degradation.
Even though the KE-Blender uses the auxiliary task in training state for enhancement, it still does not perform well in some cases.

For example, take the sentence A in Table \ref{table:cases} Case 1 as the input sentence, and the retrieved knowledge is also given, the KE-Blender outputs a better result. 
But if we change the input to sentence B of Case 1, The model indicates that it does not know the meaning of the input. To sum up, the unseen entity will have an impact on the dialogue generation model. Ignorance of unseen entities leads to a completely unavailable reply.

\begin{table}[htp!]
\centering
\caption{Demonstration of the impact of unseen entity on previous dialogue generation model.}
\resizebox{.9\textwidth}{!}{
\begin{tabular}{l|l|l}
\toprule
cases                   & entity type             & \makecell[c]{sentence}                 \\ 
\midrule
\multirow{6}{*}{\makecell[c]{\\ \\ \\Case 1}} & \multirow{3}{*}{\makecell[c]{\\ seen}}   & \begin{tabular}[c]{@{}l@{}} \textbf{Sentence A}: I have tested positive for the flu \\ so I need to go to the hospital. \end{tabular} \\ 
\cline{3-3} 
 & & \begin{tabular}[c]{@{}l@{}} \textbf{KE outputs}: \textcolor{caribbeangreen}{I hope you get it treated soon!} \\ \textcolor{caribbeangreen}{It is a viral infectious disease...} \end{tabular} \\ 
 \cline{3-3} 
 & & \textbf{GE outputs}: I hope you get well soon! I'm glad you're okay! \\
\cline{2-3} 
 & \multirow{3}{*}{\makecell[c]{\\ unseen}} & \begin{tabular}[c]{@{}l@{}} \textbf{Sentence B}: I have tested positive for the COVID-19 \\ so I need to go to the hospital. \end{tabular} \\ 
\cline{3-3} 
 & & \textbf{KE outputs}: \textcolor{red}{I have never been tested for that. What is it?} \\
\cline{3-3} 
 & & \begin{tabular}[c]{@{}l@{}}\textbf{GE outputs}: I hope it's not a serious problem. \\ I hope your symptoms are not too bad. \end{tabular} \\ 
\midrule
\multirow{6}{*}{\makecell[c]{\\ \\ \\Case 2}} & \multirow{3}{*}{\makecell[c]{\\ seen}}   & \textbf{Sentence A}: Do you like the movie Harry Potter? \\ 
\cline{3-3} 
 & & \begin{tabular}[c]{@{}l@{}}\textbf{KE outputs}: \textcolor{caribbeangreen}{I love Harry Potter!} \\ \textcolor{caribbeangreen}{I love the Harry Potter series...} \end{tabular} \\ 
\cline{3-3} 
 & & \begin{tabular}[c]{@{}l@{}}\textbf{GE outputs}: I do! I love the movies! \\ I’m a huge fan of the Harry Potter books... \end{tabular} \\ 
\cline{2-3} 
 & \multirow{3}{*}{\makecell[c]{\\ unseen}} & \textbf{Sentence B}: Do you like
the movie X-Men? \\ 
\cline{3-3} 
 & & \begin{tabular}[c]{@{}l@{}}\textbf{KE outputs}: I do! I love the X-Men movies. \\ \textcolor{red}{I love how they use a lot of xfactor characters.} \end{tabular} \\ 
\cline{3-3} 
 & & \begin{tabular}[c]{@{}l@{}}\textbf{GE outputs}: I do! I love the x X-Men movies. \\ I’m a huge fan of wolverine. \end{tabular} \\
\bottomrule
\end{tabular}}
\label{table:cases}
\end{table}

Conversely, our approach gave sentence B a good response. It captures that "\textit{COVID-19}" may be a medical related thing.

It verifies that \toolname is effective for unseen entity problems. Through the proposed method, the model captures the global co-occurrence information of entities and embeds it in the token's embedding, thereby improving the model's understanding of the current context.
\subsection{[RQ-2]: Overall Performance of \toolname}
\label{subsec:rq2}

\begin{table}[]
\centering
\caption{Comparison of our method with other state-of-the-arts.}
\begin{tabular}{ccccc}
\toprule
\multicolumn{1}{c|}{\multirow{2}{*}{Model}} & \multicolumn{2}{l|}{Test Seen}                           & \multicolumn{2}{l}{Test Unseen}     \\ \cline{2-5} 
\multicolumn{1}{c|}{}                       & \multicolumn{1}{c|}{PPL} & \multicolumn{1}{l|}{F1-score} & \multicolumn{1}{c|}{PPL} & F1-score \\ 
\midrule
\multicolumn{5}{c}{w/ knowledge during inference}                                                                                            \\ \hline
SKT + GPT-2\cite{zhao2020knowledge}         & 17.6                     & \multicolumn{1}{c|}{20.3}     & 23.7                     & 17.8     \\
SKT+PIPM+KDBTS\cite{chen-etal-2020-bridging} & 42.7                     & \multicolumn{1}{c|}{19.9}     & 65.7                     & 17.6     \\
KnowledGPT\cite{zhao2020knowledge}          & \textbf{19.2}            & \multicolumn{1}{c|}{\textbf{22.0}} & \textbf{22.3}       & \textbf{20.5}     \\ \hline
KG-Blender\cite{cui2021knowledge}           & \textbf{13.8}            & \multicolumn{1}{c|}{\textbf{18.4}} & \textbf{16.3}       & \textbf{17.8}     \\
KE-Blender\cite{cui2021knowledge}           & 14.64                    & \multicolumn{1}{c|}{18.1}     & 16.6                     & 17.4     \\
\toolname(Ours)                           & 14.9                     & \multicolumn{1}{c|}{18.0}     & 17.1                     & 16.9     \\ \hline
\multicolumn{5}{c}{w/o knowledge during inference}                                                                                           \\ \hline
Blender-FT                                  & 17.3                     & \multicolumn{1}{c|}{15.8}     & 20.7                     & 15.1     \\
KG-Blender                                  & 18.6                     & \multicolumn{1}{c|}{15.5}     & 22.7                     & 14.7     \\
KE-Blender                                  & 16.8                     & \multicolumn{1}{c|}{16.4}     & 20.6                     & 15.1     \\
\toolname(Ours)                                & \textbf{16.5}            & \multicolumn{1}{c|}{\textbf{16.7}} & \textbf{20.0}       & \textbf{15.3}     \\ 
\bottomrule
\end{tabular}
\label{table:ppl_f1}
\end{table}

\noindent \textbf{PPL \& F1-score} We conduct experiments on some benchmarks on \datasetname. The details of the comparison are shown in Table \ref{table:ppl_f1}. For scenarios where knowledge is used during inference, KnowledGPT has the most advanced performance. Benefiting from the ability to capture knowledge, our method has better metric results in knowledge-free inference. This proves the assumption we proposed previously that the representation of the context of entities can be enhanced with the global co-occurrence information of entities.

\begin{figure}[htp]
    \centering
    \includegraphics[width=8cm]{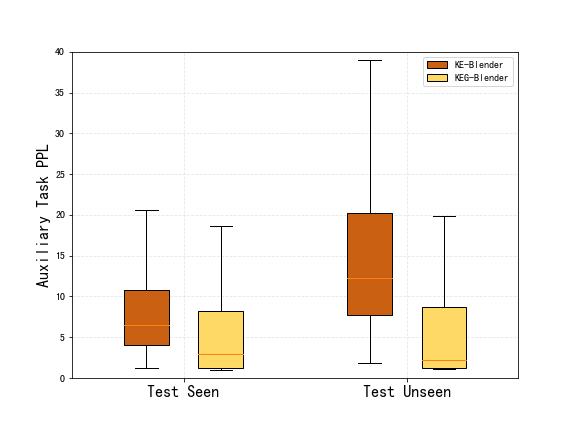}
    \caption{Boxplots of auxiliary task PPL values for KE-Blender and \toolname .}
    \label{fig:boxplot}
\end{figure}

What's more, we evaluate the auxiliary tasks of KE-Blender and \toolname with the PPL, respectively. As can be seen from Figure \ref{fig:boxplot}, on Test Seen, both approaches have smaller PPL (KE is 6.7 and GE is 4.0). On the other hand, the two approaches show a large difference in PPL (KE is 12.5 and GE is 4.9) on Test Unseen. Significantly, our auxiliary task is more effective than definition generation task on Test Unseen.

\noindent \textbf{ROUGE} Furthermore, we output the ROUGE metric to augment the comparison between KE-Blender and \toolname. As Table \ref{table:rouge} shows, our method has the best precision and recall rate on Test Unseen.

\begin{table}[ht]
\centering
\caption{Experimental results on the ROUGE metric.}
\resizebox{.8\textwidth}{!}{
\begin{tabular}{cccccccc}
\toprule
\multicolumn{4}{c|}{Rouge Precision}                                                                                                               & \multicolumn{4}{c}{Rouge Recall}                                                                                              \\ \hline
\multicolumn{8}{c}{Test Seen}                                                                                                                                                                                                                                  \\ \hline
Model                    & 1-gram                          & 2-gram                         & longest                        & Model                    & 1-gram                          & 2-gram                         & longest                         \\ \hline
Blender-FT               & 13.07                           & 1.48                           & 11.71                          & Blender-FT               & \textbf{12.84}                           & 1.57                           & 12.36                           \\
KE-Blender               & \textbf{13.53} & \textbf{1.62} & \textbf{12.25} & KE-Blender               & 13.28 & \textbf{1.73} & \textbf{12.48} \\
\toolname              & 13.29                           & 1.60                           & 11.97                           & \toolname              & 12.35                           & 1.72                           & 11.27                           \\ \hline
\multicolumn{8}{c}{Test Unseen}                                                                                                                                                                                                                                 \\ \hline
Model                    & 1-gram                          & 2-gram                         & longest                        & Model                    & 1-gram                          & 2-gram                         & longest                         \\ \hline
Blender-FT               & 12.39                           & 1.44                           & 11.24                          & Blender-FT               & 11.31                           & 1.46                           & 10.32                           \\
KE-Blender               & 12.79                           & 1.56                           & 11.63                          & KE-Blender               & 11.83                           & 1.63                           & 10.82                           \\
\toolname & \textbf{13.13} & \textbf{1.58} & \textbf{11.92} & \toolname & \textbf{12.26} & \textbf{1.67} & \textbf{11.19} \\
\bottomrule
\end{tabular}}
\label{table:rouge}
\end{table}

\textbf{Ablation Study} We studied the contribution of tasks \moduleA and \moduleB to knowledge embedding, that is, the correlation with the improvement of evaluation metrics. From the result in Table \ref{table:ablation}, it is obvious that \moduleA brings more performance improvements, which verifies our assumption that global co-occurrence information can improve the representation of the context of entities. In addition, the \moduleB also brings little improvement. It means that letting the model predict named entity tags will also brings little benefit to the model to a certain extent. In general, the combination of the two tasks brings the highest performance improvement.

\begin{table}[hbp!]
\centering
\caption{Ablation study result on Test Unseen.}
\begin{tabular}{p{3cm}p{1.5cm}rp{1.5cm}}
\toprule
Model       & PPL & F1-score \\ \hline
\toolname & \textbf{20.0}  & \textbf{15.27}   \\
w/o NER     & 20.2  & 15.22    \\
w/o Graph   & 20.6   & 15.13    \\
Blender-FT  & 20.7   & 15.08    \\
\bottomrule
\end{tabular}
\label{table:ablation}
\end{table}
\subsection{[RQ-3]: Case Study}
\label{subsec:rq3}

We have selected some examples to show the results generated by the previous method and our method. Table \ref{table:case_study} shows the dialogue generation effect of each model in two different scenarios, Seen and Unseen, respectively.

Take the first case of Table \ref{table:case_study} as an example, the entity "\textit{Harry Potter}" is general in the training set, so when generating dialogue, the model has learned enough knowledge about the context of "\textit{Harry Potter}". In the fourth case of Table \ref{table:case_study}, the entity "\textit{Grimm's Fairy Tales}" is infrequent, so previous methods do not adequately incorporate knowledge of this entity into contextual sentences. However, our model generated a relatively good response. It shows that our approach can effectively perform contextual representation enhancement based on 1-hop nodes of entities.

\begin{table}[htp!]
\centering
\caption{The output of our approach and other state-of-the-arts on selected samples.}
\resizebox{0.9\textwidth}{!}{\begin{tabular}{|cl|}
\toprule
\hline
\multicolumn{1}{|c}{\textbf{Sentence}}                                            & \multicolumn{1}{|c|}{\textbf{Output}}                                                                                                                                                                                    \\ \hline
\multicolumn{2}{|c|}{\textbf{Seen}}                                                                                                                                                                                                                                                                 \\ \hline
\multicolumn{1}{|c|}{\multirow{3}{*}{\makecell[c]{\\ \\ \\Who wrote Harry Potter?}}}            & \begin{tabular}[|c]{@{}l@{}}\textbf{Blender-FT:}\\ I'm not sure, but I do know that the series has been around since \\ the late 18th century.\end{tabular}                                                              \\ \cline{2-2} 
\multicolumn{1}{|c|}{}                                                    & \begin{tabular}[|c]{@{}l@{}}\textbf{KE-blender:}\\ I don't know who the author is, but I do know that the Harry Potter \\ series is the most popular series in the world.\end{tabular}                                   \\ \cline{2-2} 
\multicolumn{1}{|c|}{}                                                    & \begin{tabular}[|c]{@{}l@{}}\textbf{\toolname :}\\ I'm not sure, but I know that the first book was published in 1996. \\ It was written by J K. Rowling.\end{tabular}                                                  \\ \hline
\multicolumn{1}{|c|}{\multirow{3}{*}{\makecell[c]{\\ \\ \\I want to go to the Paris.}}}         & \begin{tabular}[c]{@{}l@{}}\textbf{Blender-FT:}\\ I've never been to Paris, but I'd love to go to the Museum of Art.\end{tabular}                                                                                       \\ \cline{2-2} 
\multicolumn{1}{|c|}{}                                                    & \begin{tabular}[c]{@{}l@{}}\textbf{KE-blender:}\\ I've never been there, but I'd love to go. it's the world's largest \\ art museum, and the largest in the world by area.\end{tabular}                                 \\ \cline{2-2} 
\multicolumn{1}{|c|}{}                                                    & \begin{tabular}[c]{@{}l@{}}\textbf{\toolname :}\\ I have never been there, but I would love to go to Paris, \\ the capital of France.\end{tabular}                                                                     \\ \hline
\multicolumn{2}{|c|}{\textbf{Unseen}}                                                                                                                                                                                                                                                               \\ \hline
\multicolumn{1}{|l|}{\multirow{3}{*}{\makecell[c]{\\ \\ \\Who wrote Grimm's Fairy Tales?}}}     & \begin{tabular}[c]{@{}l@{}}\textbf{Blender-FT:}\\ I'm not sure, but I do know that it was written by Georger. Martin.\end{tabular}                                                                                      \\ \cline{2-2} 
\multicolumn{1}{|l|}{}                                                    & \begin{tabular}[c]{@{}l@{}}\textbf{KE-Blender:}\\ I'm not sure, but I do know that he wrote the first volume of the series, \\ the winds of winter\end{tabular}                                                         \\ \cline{2-2} 
\multicolumn{1}{|l|}{}                                                    & \begin{tabular}[c]{@{}l@{}}\textbf{\toolname:}\\ I'm not sure, but I know that the story was written by William C. Grimm.\end{tabular}                                                                                \\ \hline
\multicolumn{1}{|l|}{\multirow{3}{*}{\makecell[c]{\\ \\ \\I want to go to the Summer Palace.}}} & \begin{tabular}[c]{@{}l@{}}\textbf{Blender-FT:}\\ I've never been to the palace, but I'd love to go to the Museum of Natural History.\end{tabular}                                                                      \\ \cline{2-2} 
\multicolumn{1}{|l|}{}                                                    & \begin{tabular}[c]{@{}l@{}}\textbf{KE-Blender:}\\ I love the palace, it is a palace of the Summer Palace, a palace that is built to serve \\ as a palace for the people of the United Kingdom and Ireland.\end{tabular} \\ \cline{2-2} 
\multicolumn{1}{|l|}{}                                                    & \begin{tabular}[c]{@{}l@{}}\textbf{\toolname:}\\ Ours \& I've never been to the Summer Palace, but I'd love to go to the Palace of Wales.\end{tabular}                                                                \\ \hline
\end{tabular}}
\label{table:case_study}
\end{table}


In Table \ref{table:train_data}, we show the training data of two modules. First of all, in \moduleA, we can see that the 1-hop node of "\textit{fiction}" taken from the graph has a strong correlation with "\textit{fiction}". In Table \ref{table:train_data}, we show the training data of two modules, including input sequence and target sequence.

\begin{table*}[htp]
\centering
\caption{Enhancement data in training stage.}
\resizebox{0.8\textwidth}{!}{
\begin{tabular}{p{2cm}p{10cm}}
\hline
\multicolumn{2}{c}{\textbf{\moduleA}}  \\ \hline
input      & It blends science \$fiction\$ and paranormal/psychological/MK Ultra type stuff together, but it's science \$fiction\$ at its core.  \\
target     & [GRAPH] genre stories fantasy author science horror novels novel star variety writers pulp doctor franchise trek elements passaic publishing plot sci [END]  \\ \hline
input      & What is your favorite thing to do with internet access? I like being able to use my computer and \$smartphone\$ to use my email and browse the world wide web.  \\
target     & [GRAPH] smartwatch browser tablet watch iphone pro ilife homepod xcode iwork suites safari macos ipad ipod multinational itunes productivity logic creativity [END]  \\ \hline
\multicolumn{2}{c}{\textbf{\moduleB}}  \\ \hline
sentence   & I think the most famous tennis match was between Bobby Riggs \& Billie Jean King.   \\
target & [NER] I think the most famous tennis match was between PERSON \& PERSON. [END]   \\ 
\hline
sentence   & PepsiCo is located in NY - I live in Georgia so we live on Coke down here.   \\
target & [NER] ORG is located in NY - I live in GPE so we live on ORG down here. [END]  \\ 
\hline
sentence   & No it is a wide grouping and I believe it originated in Western South America.   \\
target & [NER] No it is a wide grouping and I believe it originated in LOC. [END] \\
\hline
\end{tabular}}
\label{table:train_data}
\end{table*}

\section{Discussion}

\textbf{Open-domain dialogue generation still has huge opportunities for exploration.} Our model only uses 1-hop nodes for knowledge enhancement. Of course, GNN \cite{scarselli2008graph} models with stronger representation capabilities can also be used for knowledge enhancement, such as GAT \cite{velivckovic2017graph} and GCN \cite{kipf2016semi}. We only use the 1-hop node as a similar node. In fact, we can also use the node2vec method to train the node embedding, so as to use the embedding to calculate the node. Alternatively, we can also use 
PageRank \cite{page1999pagerank} to get node similarity. 
In the \moduleB stage, we only use the fine-grained Bi-LSTM+CRF model as our NER model, and we similarly use more advanced transformer-based methods such as RoBERTa \cite{liu2019roberta} to improve the model effect.

\noindent \textbf{Applicability of graph knowledge enhancement in other NLP tasks.} We can apply graph knowledge enhancement to other applications of NLP. For example, in weakly supervised text classification\cite{meng2018weakly}, we have also done a simple experiment. It has been proved by experiments that the nodes similar to the label name extracted through the graph will indeed promote the classification effect.

\noindent \textbf{Difficulty of Open-domain dialogue generation.} Open domain dialogue dataset in languages besides English and Chinese are difficult to find \cite{kann2022open}. Users usually do not have any specific goals, and the content of the dialogue is more casual, which makes the model need to guess the meaning of the user's dialogue. At the same time, due to the novelty of the conversation topic, there will be a large number of unseen entities, and some unseen entities are not associated with any entities in the historical data.

\section{Conclusion}

In this paper, we propose to use graph knowledge as a method of knowledge enhancement. We use the graph node generation task to enhance the model learning knowledge, and use the named entity tags to expand unseen entities.
Experiments show that the proposed approach works well on unseen entity problems. Our method does not require additional input of knowledge information in the inference stage, which provides great convenience and usability for inference work.

\bibliographystyle{splncs04}
\bibliography{mybib.bib}

\end{document}